\documentclass[preprint, numafflabel, 12pt]{elsarticle}
\usepackage{graphicx}
\usepackage{float}
\usepackage{booktabs}
\usepackage{doi}

\usepackage[linesnumbered,ruled,vlined, algo2e]{algorithm2e}
\usepackage[utf8]{inputenc}
\usepackage{float}
\usepackage[section]{placeins}
\usepackage{graphicx}
\usepackage{subcaption}

\usepackage{amssymb}
\usepackage{amsmath}

 \usepackage{lineno}

\journal{Internet of Things}

\begin{document}

    \begin{frontmatter}

   \title{Multi-Agent System-driven Digital Twins for predictive maintenance: architectures, technologies and open research challenges}
    
    \author[1]{Korota Arsène COULIBALY}
    \ead{korotaarsne.coulibaly@usmba.ac.ma}
    
    \author[1]{Mohamed HAMLICH}
    
    \cortext[cor1]{Corresponding author.}
    
    \affiliation[1]{organization={LCCPS Lab, ENSAM, Hassan II University of Casablanca},
                addressline={150 Bd du Nil},
                city={Casablanca},
                postcode={20670}, 
                state={},
                country={Morocco}}

    \begin{abstract}
        Digital twins have emerged as a foundational technology within the context of the Fourth Industry 4.0, offering a paradigm for the real-time virtual representation of physical systems. However, managing their growing complexity, particularly in distributed industrial environments, requires intelligent architectures capable of autonomous decision-making, dynamic adaptability, and inter-agent coordination. This systematic review explores the intersection between Multi-Agent Systems and Digital Twins, with a particular focus on predictive maintenance applications in resource-constrained contexts. Through a critical analysis of over 547 papers published in high-impact journals (IEEE Transactions, Nature, Elsevier, MDPI), we establish a taxonomy of existing hybrid architectures, identify persistent technological bottlenecks, and formulate three open research questions concerning: (i) the deployment of artificial intelligence on resource-constrained microcontrollers, (ii) distributed multi-node coordination via lightweight communication protocols, and (iii) the hierarchical orchestration of Digital Twins toward smart factory control integrating residual life estimation and explainable Artificial Intelligence. The results of this analysis reveal that, despite significant progress, no existing system offers an integrated embedded-distributed hierarchical solution that simultaneously meets the requirements of Industry 5.0.
    
    \end{abstract}

    
    
    \end{frontmatter}
    
    

    \section{Introduction}
    
        The digital transformation of industrial production systems has accelerated significantly since the advent of Industry 4.0, driven by the convergence of Internet of Things (IoT) technologies, cloud computing, Artificial Intelligence(AI), and low latency wireless communications\cite{MEL26},\cite{YAN23a},\cite{YUA24},\cite{ZHU23}. In this context, the concept of the Digital Twin (DT) has emerged as a unifying paradigm, enabling the creation of a dynamic and synchronized virtual representation of a physical system, its behaviors, and its state history \cite{TAO19a},\cite{TAO18a}. The foundational definition, attributed to Grieves in 2014 \cite{GRIV14}, established the foundations of a conceptual trinity: physical object, virtual space, and bidirectional information flow, which has since been expanded by multidimensional frameworks, notably the five-dimensional model by Tao et al. \cite{TAO19a}, which incorporates data, services, and connections.
    
        However, the rise of DT in distributed industrial environments is revealing fundamental architectural limitations. Most current implementations rely on centralized or semi-centralized models that concentrate inference and decision-making capabilities on remote servers or in the cloud, generating latencies incompatible with the real-time requirements of industrial predictive maintenance \cite{ABD24}, \cite{LAT23},\cite{OTI25}. Furthermore, the increasing complexity of Cyber-Physical Production Systems (CPPS) requires intelligent, decentralized, and resilient management, for which Multi-Agent Systems (MAS) provide a particularly well-suited theoretical and practical framework \cite{VRA21},\cite{LOC26},\cite{WAN21}.

        In fact, MAS systems offer inherent properties that align with the needs of modern DTs: local decision-making autonomy, the emergence of coherent global behaviors from local interactions, fault tolerance through distributed redundancy, and the ability to dynamically adapt to changes in the operational environment \cite{BOW22},\cite{LAT23}. The integration of MAS into architectures therefore represents a promising approach to addressing the challenges of Industry 5.0, which places human-machine collaboration, sustainability, and resilience at the heart of its objectives \cite{REN24}. However, despite the abundance of separately published work on DTs and MAS, the literature lacks a systematic and critical review of their intersection, particularly in the context of predictive maintenance on embedded systems subject to resource and latency constraints.
        
        To address this gap, we propose an extension of the DT concept, focused on the multi-agent governance of its cognitive functions. We defines a MAS-driven DT as an architecture of virtual replicas that are nested and synchronized in real time with an embedded physical system, structured across multiple levels of abstraction (subsystem, machine, overall system), where perception, diagnostic, and prognostic functions are distributed among autonomous and cooperative agents at each level, with higher-level agents aggregating diagnostics from lower levels to detect, locate, and anticipate system failures.
        This definition serves as the conceptual foundation for this study, guiding both the inclusion criteria applied in the systematic literature review. This paper aims to fill this gap by proposing: 
                
        \begin{enumerate}[C\arabic*:]
        	\item A structured taxonomy of DT architectures driven by MAS;
        	\item A critical comparative analysis of existing approaches; 
            \item The identification of unanswered technological challenges;
            \item The formulation of open research questions aligned with the needs of the smart manufacturing industry and a proposal of a three-tier MAS-DT (Edge–Fog–Cloud) architecture that is emerging as a leading research direction
        \end{enumerate}
        
        Our paper is structured as follows: Section 2 presents the systematic review methodology adopted. Section 3 outlines the evolution of DTs and establishes their taxonomy. Section IV examines the fundamentals of MAS and their relevance to DTs. Section 5 constitutes the analytical core of the review, featuring an in-depth comparative study of hybrid DT-MAS architectures. Section 6 deals specifically with embedded predictive maintenance. Section 7 analyzes the persistent technical challenges. Section 8 formulates the open research questions. Section 9 presents the discussion part. Section 10 concludes with a look toward Industry 5.0.
    
    \section{Methodology}

        The main objective of this systematic review is to answer the following research question: What reference architecture emerges from the literature regarding the convergence of DTs and MAS as applied to edge predictive maintenance, and what methodological limitations currently affect the published research? This question directly guided the literature search strategy and the inclusion and exclusion criteria. 
        This review follows the PRISMA (Preferred Reporting Items for Systematic Reviews and Meta-Analyses) protocol, adapted for literature reviews in computer engineering and embedded systems. The literature search strategy targeted the following databases: IEEE Xplore, Scopus, Web of Science, ScienceDirect (Elsevier), SpringerLink, and MDPI. The literature search relied on a combination of controlled keywords and Boolean operators, organized around two thematic areas corresponding to the core contributions of this study: 
        \begin{itemize}
            \item Digital Twins: "digital twin" OR "cyber-physical systems" OR "virtual model"
            \item MAS and embedded intelligence: "multi-agent systems" OR "Edge AI" OR "edge computing", "TinyML" OR "predictive maintenance".
        \end{itemize}
        The period covered ranges from 2014, the publication year of Grieves seminal work, to early 2026.
        
\begin{figure}[htpb]
            \centering
                \includegraphics[width=\textwidth]{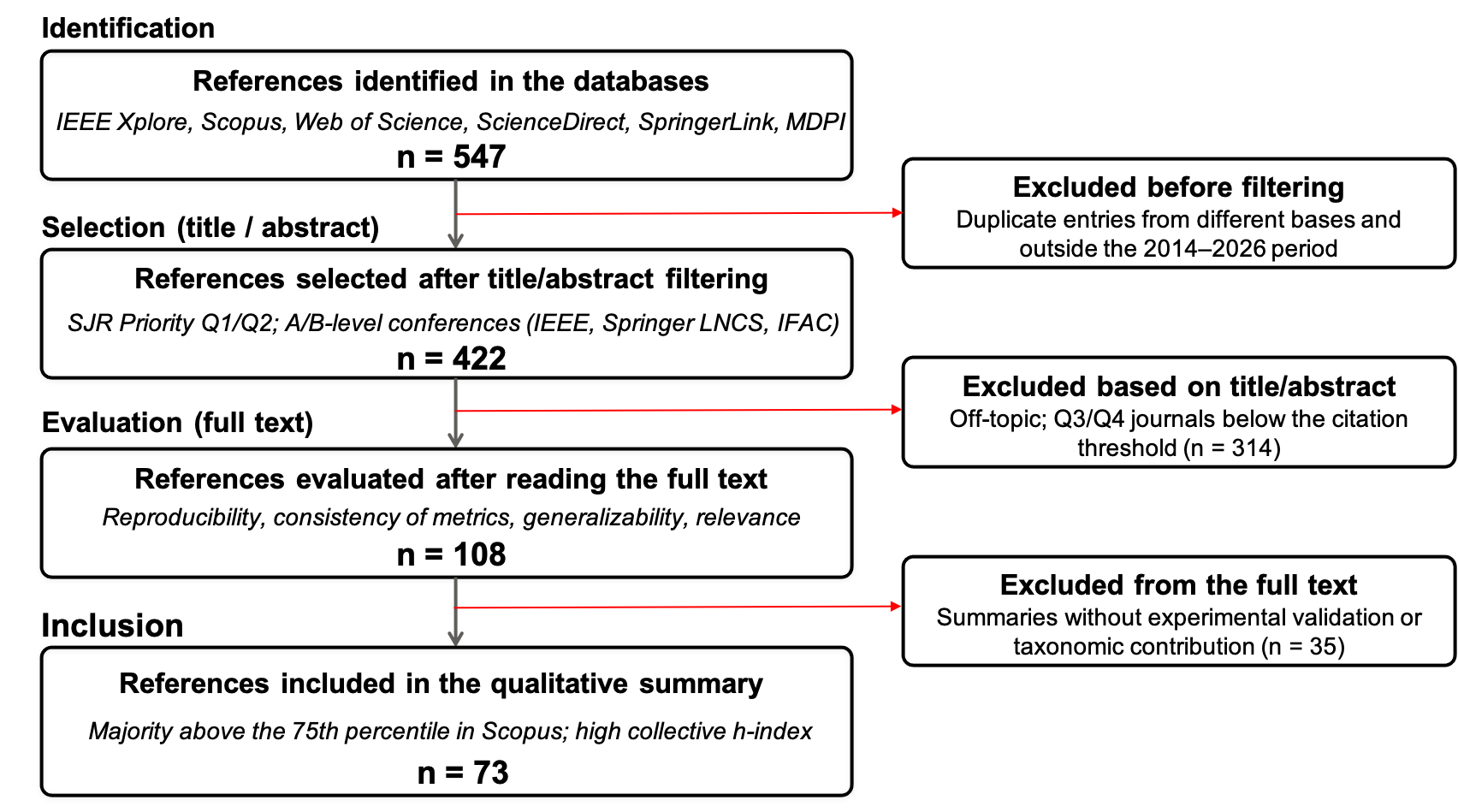}
                \vspace{.7em}
                \caption{PRISMA flowchart}\label{fig:ref_workflow}
            \end{figure}

        The eligibility criteria prioritized articles published in journals indexed in the first or second quartile according to the SJR (Scimago Journal Ranking), with a minimum of five citations per year for publications prior to 2022. Conference papers were included if they came from A- or B-level events (IEEE, Springer LNCS, IFAC) and presented original experimental or architectural contributions. General review articles without experimental validation were excluded unless they presented a significant taxonomic or methodological contribution.
        In total, 547 initial references were identified, of which 422 were retained after filtering by title and abstract, and 73 were retained after full reading and evaluation of scientific quality as shown in Fig. \ref{fig:ref_workflow}. The quality criteria used included: the reproducibility of experiments, the rigor of evaluation metrics, the generalizability of conclusions, and relevance to the targeted issues. The 63 selected references have a high collective h-index, with the majority ranked above the 75th percentile in their respective fields according to Scopus.

        The thematic analysis was then conducted using a standardized extraction grid comprising the following dimensions: architecture type (centralized, distributed, hierarchical, hybrid), deployment level (cloud, fog/edge, embedded), communication protocol used, integrated Artificial Intelligence method, target application, and reported performance metrics. This framework was used to construct the comparative tables presented in the following sections.

    \section{Evolution and Taxonomy of Digital Twins}
    
    \subsection{Genesis and conceptual Evolution}
    
\begin{figure}[htpb]
                \centering
                \includegraphics[width=\textwidth]{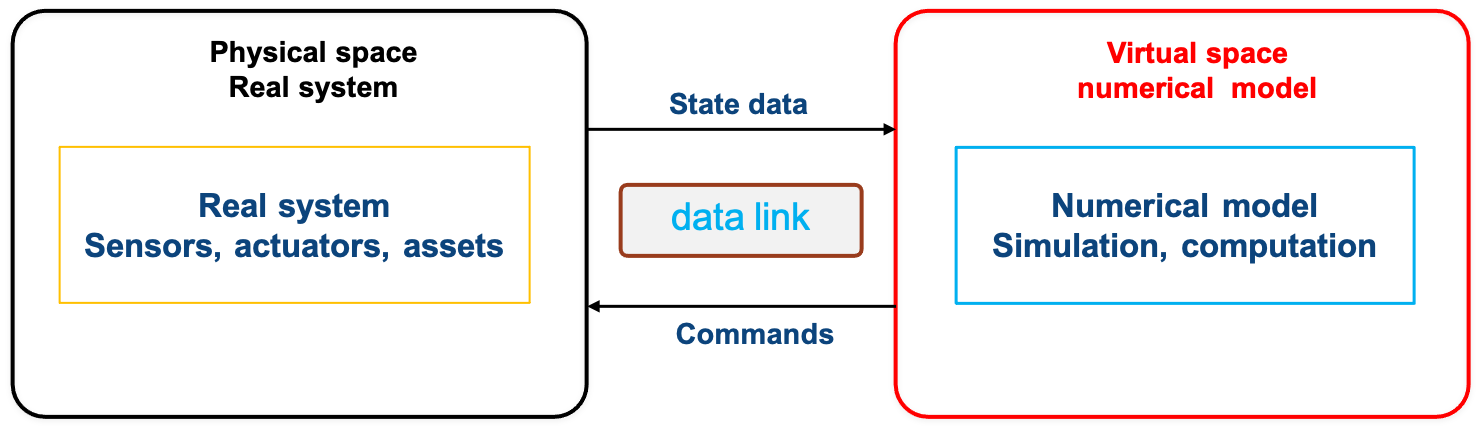}
                \caption{Grieves Digital architectures}\label{fig:grieves_dt}
            \end{figure}

        The concept of the DT has its roots in product lifecycle modeling work developed in the U.S. aerospace industry in the early 2000s, before being formalized by Grieves \cite{GRIV14} under the term "Digital Twin" within the context of product lifecycle management (PLM). The initial formulation identified three essential components: the physical space (the real system), the virtual space (the numerical model), and the data link connecting the two. This three-part framework, though influential, proved insufficient to capture the richness of dynamic interactions in modern Cyber Physical Systems (CPS).  Fig. \ref{fig:grieves_dt} shows GrieGrieves'hitecture.

\begin{figure}[htpb]
                \centering
                \includegraphics[width=\textwidth]{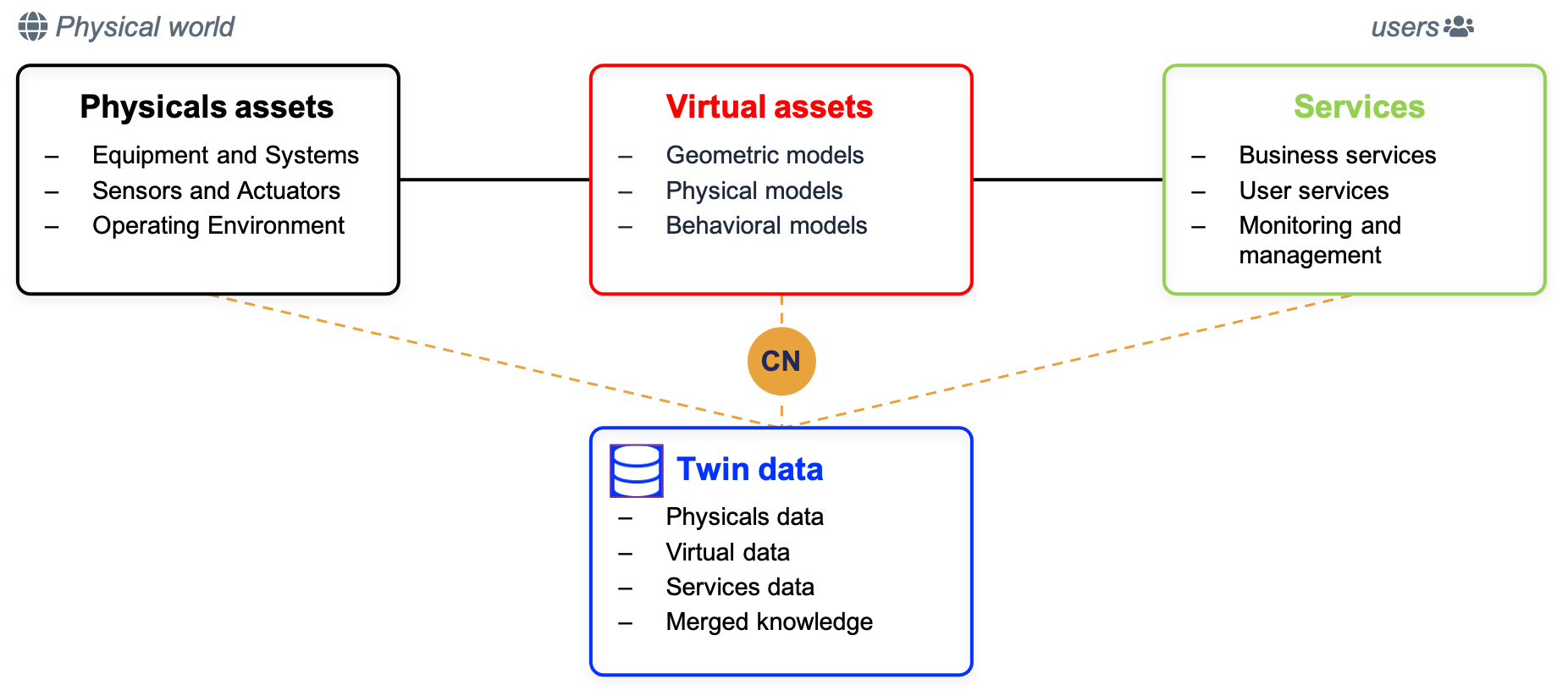}
                \caption{Digital twins are the core of the model: they connect physical assets, virtual assets, and services in real time through connections.
                }\label{fig:tao_dt}
            \end{figure}

        It was Tao et al. \cite{TAO19b},\cite{TAO18b},\cite{QIQ21} who gradually enriched this framework by proposing the five-dimensional model (5D-DT) as shown in Figure \ref{fig:tao_dt}, which integrates: physical assets or assets, virtual entities, twin data, services, and connections. This model has served as a reference for the scientific community, accumulating over 2000 citations in Scopus, and has guided the development of increasingly sophisticated architectures. The review by Liu et al. \cite{LIU23a} subsequently systematized this evolution by distinguishing four orthogonal axes of characterization: physical entities, virtual models, twin data, and applications.

        The evolution of DTs can be divided into four distinct phases. The first phase (2014-2017) corresponds to the emergence of the concept, dominated by aerospace applications and academic prototypes \cite{RIO15},\cite{HRI13}. The second phase (2017-2020) marked the expansion into the manufacturing industry, with seminal work such as that of Tao \cite{TAO17},\cite{TAO18c} which defined the digital workshop paradigm. The third phase (2020-2023) is characterized by application diversification: energy, healthcare, agriculture, smart cities, and by the growing integration of Artificial Intelligence \cite{ZHO23},\cite{CHE23},\cite{KOB24}. The fourth phase, currently underway, is marked by architectural decentralization, the deployment of intelligence at the edge, and convergence with MAS to meet the requirements of Industry 5.0 \cite{ABD24}\cite{VRA21},\cite{LAT23},\cite{YOU26}.

    \subsection{Multidimensional taxonomy}

\begin{figure}[htpb]
                \centering
                \includegraphics[width=\textwidth]{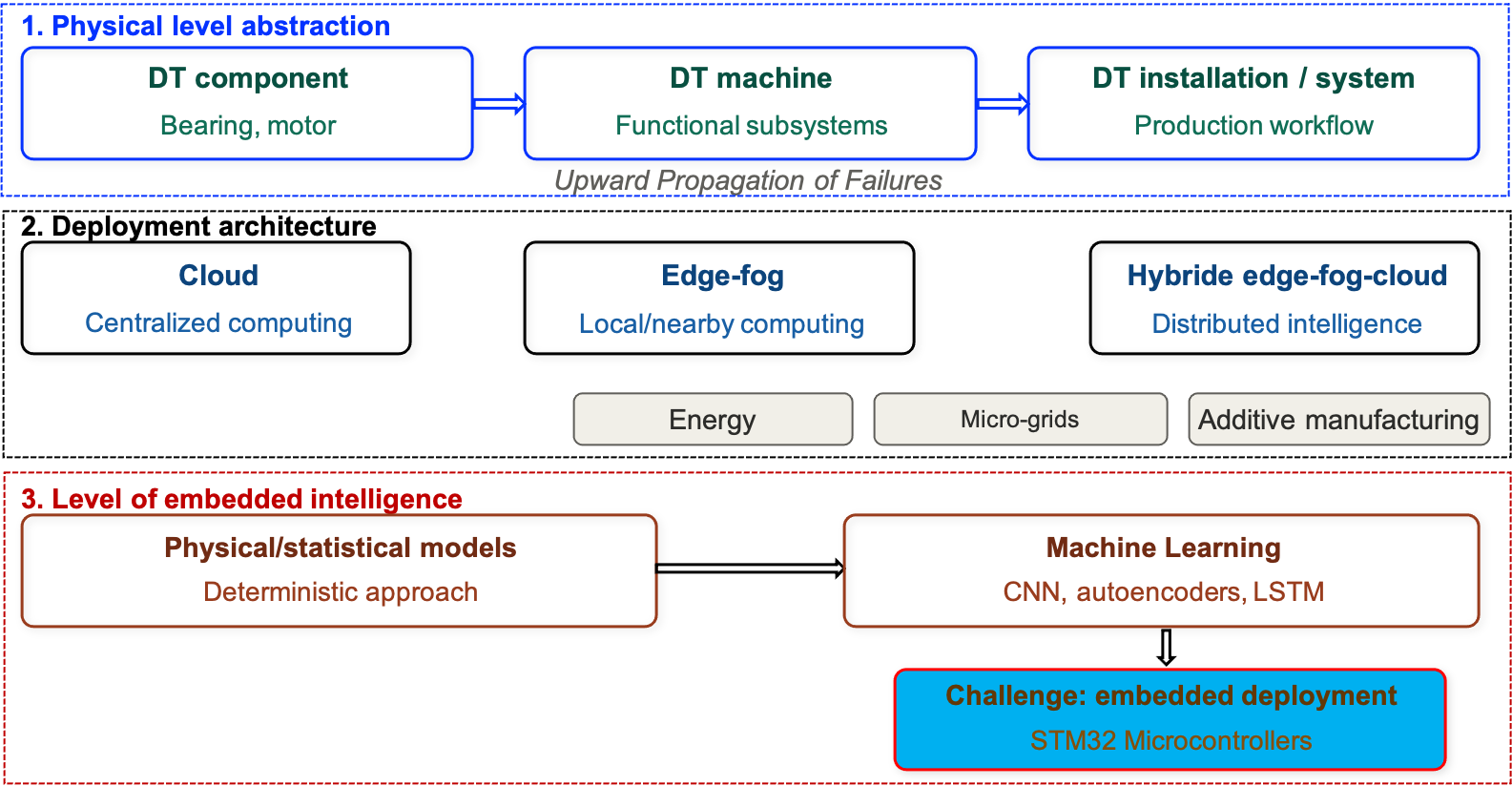}
                \caption{A total of 5 dimensions: physical abstraction, deployment, embedded intelligence, temporal granularity, and interaction mode}\label{fig:taxonomy_dt}
            \end{figure}

    The literature proposes several orthogonal criteria for classifying DTs architectures. A comprehensive taxonomy must include at least five dimensions: (1) the level of physical abstraction, (2) the deployment architecture, (3) the degree of embedded intelligence, (4) the temporal granularity, and (5) the mode of interaction with the physical system(see Fig. \ref{fig:taxonomy_dt}).
        
    Regarding the level of physical abstraction, we distinguish between component-level DTs, which model an elementary component such as a bearing or motor, machine-level DTs, which aggregate multiple components into functional subsystems, and plant or system-level DTs, which orchestrate multiple machines within a coherent production flow \cite{BOT23},\cite{ZHO23},\cite{LAT23}. This hierarchy reflects an industrial reality in which failures often propagate from components to systems, justifying a bottom-up modeling approach.
        Deployment architecture constitutes a second key focus area. Julien and Martin \cite{JUL21} propose a use-centric classification that distinguishes between cloud-centric architectures, edge-fog architectures, and hybrid architectures. The work of Qi and Tao \cite{QIQ19} laid the groundwork for edge-fog-cloud integration in DT architectures, demonstrating that intelligence must be distributed across all three layers to optimize the latency, bandwidth, and computing power trade-off. Concrete implementations have validated this approach in fields as diverse as energy \cite{YU22}, microgrids \cite{BAZ22}, and additive manufacturing \cite{JYE23}.
        
        The degree of embedded intelligence is becoming an increasingly critical differentiator. Early DTs architectures relied on simple deterministic or statistical physical models. The integration of Machine Learning in particular Convolutional Neural Networks (CNNs), autoencoders, and Long Short-Term Memory (LSTM) models, has transformed DTs into learning and predictive entities \cite{CHE23},\cite{KOB24},\cite{ZHO23}. However, deploying these models directly on resource-constrained embedded systems remains an open challenge, as evidenced by the limited number of studies documenting implementations on STM32 type microcontrollers or similar devices.

    
    \section{Multi-Agent Systems: foundations and relevance for Digital Twins}

    \subsection{Fundamental principles of Multi-Agent Systems}

        A MAS is defined as a set of autonomous agents capable of perceiving their environment, making local decisions, and interacting with other agents to achieve individual or collective goals \cite{BUR25},\cite{YOU23},\cite{VRA21},\cite{SKO23},\cite{KIM23}. The fundamental properties of an agent, autonomy, responsiveness, proactivity, and sociability align precisely with the requirements of a distributed DT architecture. Autonomy ensures operational continuity even in the event of a network failure. Reactivity ensures a rapid response to physical events. Proactivity enables the initiation of preventive actions based on predictions. Sociability facilitates inter-twin coordination for orchestrated maintenance at the plant level\cite{XUC23}.
        Agent interaction protocols constitute a critical dimension that is often underestimated in the literature on DTs. The Contract Net Protocol (CNP), stemming from the seminal work of Smith (1980)\cite{SMIT80} and widely studied since, remains the benchmark for task allocation and negotiation mechanisms in industrial MAS. Its adaptation to the constraints of embedded environments, particularly regarding computational complexity and network bandwidth, represents a significant technological bottleneck. Minor adaptations of the CNP have been proposed for IoT environments, but their validation on microcontrollers with limited memory remains limited in the literature\cite{LAT23},\cite{VRA21}.
        Holonic agent architectures, which stem from Koestler's work, have been applied in specific contexts within intelligent manufacturing systems. The Agent-Robot-Task-Interface architecture, proposed in \cite{BOR19}, represents one of the first attempts to integrate embedded DTs into a holonic framework, demonstrating the feasibility of a recursive organization in which each node can simultaneously act as an agent in an MAS and as the DT of a physical component. This duality represents a major architectural lever for future developments.

    \subsection{Multi-Agent Systems and Digital Twin coupling: state of the art}
    
        The convergence of MAS and DTs has given rise to several hybrid architectures documented in the literature. Vrabic et al. \cite{VRA21} proposed an intelligent agent architecture for DT resilience in manufacturing, demonstrating that agents can detect and compensate for degradation in the synchronization quality between the virtual model and the physical system. Latsoú et al. \cite{LAT23} extended this approach to anomaly detection and bottleneck identification in complex production systems, with experimental validation on a real assembly line. This line of work aligns with the connectivity topologies formalized by Schroeder et al. \cite{SCH21}, which distinguish star, mesh, and hierarchical architectures for networking DTs, and finds a methodological extension in Qamsane et al. \cite{QAM21}. They propose a structured methodology for deploying DT solutions for manufacturing systems.
        
        Wan et al. \cite{WAN21} proposed a recursive multi-agent DT model for energy management in connected materials, introducing the concept of a recursive DT in which an agent can itself be modeled by a nested DT. This recursive approach opens up significant architectural possibilities for modeling complex hierarchical systems, but raises unresolved questions about managing data consistency across levels of abstraction. Questions that Hamzaoui and Julien \cite{HAM22} address from the perspective of the social ecosystems of networked DTs, and that \cite{AZI23},\cite{KAM22},\cite{BLA22} formalize by proposing distributed DTs as proxies that offer composability and flexibility rather than a rigid hierarchy of nested models.
        
        Villalonga et al. \cite{VIL20} studied local decision-making within a distributed DT framework, highlighting that decentralizing reasoning to physical nodes significantly reduces reliance on a robust cloud infrastructure and improves the system's overall resilience.
        Borangiu and his co-authors \cite{BOR19},\cite{BOR20} documented several production control implementations integrating embedded DTs into MAS type architectures, confirming the industrial feasibility of this approach in semi-continuous production contexts. This feasibility is corroborated by Caesar et al. \cite{CAE23}, whose DT reconfiguration management framework precisely addresses the need for agent topologies capable of dynamically evolving in a production environment, and by Houé et al. \cite{HOU23}, who demonstrate the viability of a modular and distributed architecture integrating an embedded DT for adapting assistive technologies. Further upstream, Jazdi et al. \cite{JAZ21} situate these implementations within the broader perspective of integrating AI into industrial automation systems driven by intelligent DTs. However, Tripathi et al. \cite{TRI24} emphasize that effective collaboration between stakeholders remains one of the major emerging challenges for consolidating such distributed digital twin ecosystems.
        Despite these advances, a critical analysis of the literature reveals persistent gaps:
        
        \begin{enumerate}[a)]
        	\item First, the majority of proposed architectures run agents and DT models on servers or desktop computers, rather than on embedded microcontrollers.
        	\item Second, coordination mechanisms between agents are rarely validated under real-world network conditions with bandwidth constraints.
            \item Third, the integration of deep learning models into agents remains problematic in terms of interpretability and operator confidence.
        \end{enumerate}

    
    \section{A Comparative analysis of hybrid Digital Twins and Multi-Agent System architectures}

        \subsection{Classification of existing architectures}
        
        A review of the literature identifies five major architectural families for DTs driven by intelligent agents (see Table 1). These families are distinguished primarily by their deployment topology, their inter-agent coordination mechanisms, and their level of decision-making autonomy.
        The first family, which we refer to as enriched client-server architectures, corresponds to approaches in which a central server hosts the virtual models and one or more peripheral agents collect and transmit data. Although simple to implement, these architectures are highly dependent on network availability and pose a single point of failure. The work of Kherbache et al. \cite{KHE22},\cite{KHE23} on Eclipse Ditto and Hono exemplifies this family, with feasibility demonstrations in IIoT contexts, but without the integration of advanced multi-agent coordination mechanisms.
        The second category comprises distributed fog-edge architectures, in which intermediate computing nodes host partial DTs and local decision-making agents. These architectures, advocated notably by Qi and Tao \cite{QIQ19} and documented in \cite{ABD24}, offer a better latency-availability trade-off. Synchronization between DTs at different levels constitutes their main technical challenge.
        The third family, holonic and recursive architectures, relies on the principles of holonic MAS to organize DTs into dynamic hierarchies. \cite{BOR19},\cite{WAN21} represent the most advanced contributions in this direction, although their experimental validation remains limited to controlled laboratory environments.
        The fourth category, architectures based on embedded autonomous agents, aims to deploy intelligent agents directly on processing units within physical devices. This category is the least documented in the literature despite its considerable practical value. \cite{HIN19} laid the conceptual groundwork for using open-source microcontrollers in DT communication, but did not address the issue of constrained local inference.
        The fifth, emerging family corresponds to industrial metaverse architectures in which DTs are integrated into mixed-reality environments to facilitate human-machine interaction \cite{REN24}\cite{YOU26}. These architectures, while promising for Industry 5.0, still exhibit a limited level of technological maturity for large scale industrial deployments.

        \begin{table}[htbp]
        \centering
        \caption{Comparison of Digital Twin architectures driven by intelligent agents.}
        \label{table_2} 
        \resizebox{\textwidth}{!}{
            \begin{tabular}{l l l l l l l l}
                \toprule
                References & Architectures type & Deployment level & Comm. Protocol & AI method & Application & Accuracy/efficiency & Validation \\ 
                \midrule
                \cite{VRA21} & Resilient intelligent agent & cloud/edge & REST/HTTP & rules+ML & manufacturing & Improved resilience & experimental \\
                \cite{LAT23} & distributed Multi-Agent & cloud/edge & OPC-UA & CNN + SVM & Bottleneck detection & $>$92\% accuracy & experimental \\
                \cite{WAN21} & Recursive multi-agent DT & edge/fog & MQTT & fuzzy logic & Energy management & 18\% energy savings & Simulation \\
                \cite{BOR19} & Embedded Holonic ARTI & embedded edge & owner & Business rules & Semi-continuous production & response time $<$50ms & Industrial \\
                \cite{BOR20} & CLOUD + embedded DT & cloud+edge & OPC-UA & adaptive thresholds & Smart Manufacturing & Uptime $>$99\% & Industrial pilot \\
                \cite{OUA21} & distributed edge-cloud & cloud+edge & MQTT & data processing & Smart Manufacturing & Latence $<$100ms & Prototype \\
                \cite{ABD24} & distributed IIoT & edge/fog/cloud & MQTT+REST & LSTM+RF & IIoT predictive maintenance & MSE down 34\% & simulation \\
                \cite{KHE22} & Eclipse Ditto + Hono & edge/cloud & MQTT/AMQP & rules & IIoT network & Proven interoperability & prototype \\
                \cite{VIL20} & local distributed DT & edge & OPC-UA & local reasoning & CPS production & Latence $<$30ms & experimental \\
                \cite{SKO20} & MAS + DTs Plants & edge+cloud & owner & standard MAS & agriculture CPS & accuracy 87\% & pilot \\
                \cite{HIN19} & MCU open-source DT & embedded & serial/WiFi & No local AI & Smart Manufacturing & real-time synchronization & prototype \\
                \cite{LIS23} & DT focal modulation & cloud & REST & CNN focal modulation & fault diagnosis & accuracy $>$90\% & experimental \\
                \cite{KOB24} & DT explainable RUL & cloud & REST/API & LSTM + SHAP & RUL estimation & error $<$8\% & experimental \\
                \cite{ZHO23} & predictive maintenance DT & cloud & REST & LSTM + CNN & application review & survey & review \\
                \cite{CHE23} & ML maintenance DT & cloud/edge & REST & ML multiple & manufacturing & survey & review \\
                \bottomrule
            \end{tabular}
            }
        \end{table}
    
    \subsection{A critical analysis of communication mechanisms}

        Communication between agents forms the backbone of any MAS, and adapting it to the constraints of IIoT environments represents a major technical challenge. Communication protocols can be classified according to three orthogonal criteria: topology (point-to-point, publish-subscribe, broadcast), message format (verbose/JSON, binary/Protobuf, proprietary), and delivery guarantees (at-most-once, at-least-once, exactly-once).
        
        Message Queuing Telemetry Transport(MQTT), operating on a publish-subscribe model via a centralized broker, has established itself as the standard for lightweight industrial IoT communication. Its low bandwidth consumption headers do not exceed 2 bytes and its support for different levels of quality of service (QoS 0, 1, 2) makes it a protocol well suited to degraded network environments \cite{SING23},\cite{ALFU15},\cite{MISH18}. However, its reliance on a centralized broker constitutes an architectural weakness that MQTT-SN (for Sensor Networks) approaches or distributed broker implementations aim to address.
        
         Open Platform Communications Unified Architecture(OPC-UA) is the go-to solution for industrial applications requiring stronger guarantees of security and interoperability \cite{MAHN09}. Its rich information model and service discovery mechanisms make it a natural candidate for integration into digital twin architectures. However, its memory and computational footprint, several megabytes of libraries, make it difficult to deploy on STM32F4-class microcontrollers without significant optimizations.
         
        Google's Protocol Buffers (Protobuf), and in particular their Nanopb implementation for embedded systems, offer a compact and efficient binary serialization alternative for messages between agents in bandwidth-constrained environments \cite{GOOG08}. A comparison between JSON and Nanopb shows message size reductions of around 60 to 80\%, which can be critical for real-time applications on busy industrial networks. This observation motivates the exploration of Nanopb as the reference message format for embedded MAS architectures. Table 2 shows an analysis and comparison of these communication protocols.
    
    \begin{table}[htbp]
        \centering
        \caption{Comparison of communication protocols for embedded MAS-DT architectures.}
        \label{table_3}
        \resizebox{\textwidth}{!}{
            \begin{tabular}{l l l l l l l l}
                \toprule
                Protocol & Topology & Message format & Overhead (octets) & QoS & Security & MCU footprint (KB) & Suitable for Embedded\\ 
                \midrule
                MQTT v3.1.1 & pub/sub (broker) & binary/JSON & 2--5 & QoS 0, 1, 2 & optional TLS & $\sim$50--80 &  yes \\
                MQTT-SN & pub/sub (no broker) & binary & 2 & QoS 0, 1, 2 & limited & $\sim$20--40 &  highly suitable \\
                OPC-UA & client/server & binary/xml & $\sim$100--500 & high & integrated PKI & $\sim$500--2000 &  no (too heavy) \\
                CoAP & REST (UDP) & binary/CBOR & 4--20 & confirmable & DTLS & $\sim$30--60 &  partial \\
                HTTP/REST & client/server & json/xml & $\sim$200--1000 & TCP & TLS/SSL & $\sim$100--300 & prototype \\
                DDS & pub/sub (no broker) & CDR binary & $\sim$50--200 & multiple QoS & integrated & $\sim$300--500 & partial (DDS-XRCE) \\
                Nanopb (Protobuf) & serialization only & compact binary & 3--10$\times$ $<$ JSON & N/A & via transport layer & $\sim$5--20 &  highly suitable \\
                AMQP & broker message & binary & $\sim$60--200 & multiple & SASL/TLS & $\sim$200--400 &  no \\
                WebSocket & full-duplex & JSON/binary & $\sim$50--100 & TCP & WSS/TLS & $\sim$150--300 & partial \\
                CAN bus & shared bus & binary/compact & 0--8 & priority & physical only & $\sim$5--15 &  industrial \\
                SPI bus & shared bus & -- & -- & -- & physical only & $\sim$5--15 &  industrial \\
                I2C bus & shared bus & -- & -- & -- & physical only & $\sim$5--15 &  industrial \\
                \bottomrule
            \end{tabular}
        }
    \end{table}

    \subsection{Comparative analysis of deployment platforms}

    The choice of hardware platform is a critical factor that is often ignored in architectural proposals found in the literature. Most articles present architectures validated on servers or conventional computers, and then assume without proof that they can be adapted to embedded systems. This top-down approach carries significant risks of misalignment between theoretical assumptions and the physical constraints of real-world industrial deployments.
    Microcontrollers from the STM32 family (STMicroelectronics), and more specifically the STM32F429I used in recent work on embedded AI for predictive maintenance, offer a representative example of the constraints to consider: 2 MB of Flash, 256 KB of RAM, and a 180 MHz Cortex-M4 processor with a floating-point unit (FPU). These resources are theoretically insufficient for conventional deep learning models, whose memory requirements typically run into the hundreds of megabytes. However, recent advances in model quantification, pruning, and distillation have demonstrated that it is possible to deploy useful inferences in these constrained environments, provided that model architectures are designed specifically for the target hardware.
    Raspberry Pi devices and equivalent development boards (NVIDIA Jetson Nano, BeagleBone) offer a mid-range level of computing power, making them well-suited for the role of fog coordinator in a multi-tier architecture. These platforms can host JADE (Java Agent DEvelopment Framework) agents or lightweight Python implementations of negotiation protocols, while managing communication with lower-level microcontrollers and upper-level cloud servers. This three-tiered organization aligns with the trend documented in \cite{ABD24},\cite{OUA21}, which validates the edge-fog-cloud architecture for distributed DTs. Table 3 presents a comparison of the different hardware deployment platforms.
    
    \begin{table}[htbp]
    \centering
    \caption{Comparison of hardware platforms for the deployment of embedded Digital Twins.}
    \label{table_1}
    \resizebox{\textwidth}{!}{
        \begin{tabular}{l l l l l l l l l}
            \toprule
            Platforms & CPU & Flash(MB) & RAM(MB) & FPU & OS/RTOS & Connectivity & DT-MAS role & Cost(USD) \\ 
            \midrule
            STM32F429I-DISCO & Cortex-M4 @ 180MHz & 2 & 0.25 & yes & FreeRTOS/Zephyr & ETH+SPI+I2C+UART & local agent & $\sim$25 \\
            STM32H743 & Cortex-M4 @ 480MHz & 2 & 1 & DP-FPU & FreeRTOS/Zephyr & ETH+SPI+I2C+UART & advance agent & $\sim$35 \\
            ESP32 & Dual-core-xtensa @ 480MHz & 4 & 0.52 & yes & FreeRTOS/Zephyr & WiFi, BT, SPI, I2C & IoT agent & $\sim$15 \\
            Raspberry Pi 4B & Cortex-A72 @ 1.8GHz & -- & 4000 or 8000 & NEON & Linux & WiFi, BT, SPI, I2C, ETH & fog coordinator & $\sim$80 \\
            NVIDIA Jetson Nano & Cortex-A57 + GPU & -- & 4000 or 8000 & GPU 128c & Linux+CUDA & WiFi, BT, ETH & fog AI & $\sim$350 \\
            BeagleBone black & Cortex-A8 @ 1GHz & 4 & 512 & NEON & Linux(debian) & ETH+USB & light fog & $\sim$60 \\
            Arduino Uno & ATmega328 @ 16MHz & 0.032 & 0.002 & No & bare metal & serial/SPI & too limited & $\sim$15 \\
            AWS IoT greengrass & x86/ARM cloud & -- & $>$1000 & No & Linux & cloud AWS & cloud factory & SaaS \\
            Azure IoT edge & x86/ARM cloud & -- & $>$1000 & No & Linux/Windows & Azure cloud & cloud factory & SaaS \\
            \bottomrule
        \end{tabular}
        }
    \end{table}
    
    \section{Embedded predictive maintenance: AI-Digital Twins and Multi-Agent Systems convergence}

    \subsection{Specific features of predictive maintenance for embedded systems}
    
        Predictive maintenance based on DTs is one of the most mature and well documented applications in the literature \cite{ZHO23},\cite{CHE23},\cite{TAO18c}. The convergence of three technological trends high-resolution sensors, machine learning models, and synchronized virtual representations has enabled the development of systems capable of predicting failures with unprecedented accuracy and lead time. However, nearly all documented architectures offload inference calculations to remote servers, creating a dependency on the network that can be critical in industrial environments where connectivity is not guaranteed.
        The specific challenges of embedded predictive maintenance fall into three categories. First, computational constraints: conventional machine learning algorithms, deep neural networks, random forests, support vector machines involve computational complexities and memory footprints that are incompatible with industrial-grade microcontrollers. Second, time constraints: predictive maintenance requires a response within a limited timeframe, with latencies typically under 100 ms for applications detecting faults during the process. Third, the power constraint: embedded nodes in industrial environments are often battery-powered or rely on energy harvesting, imposing constraints on the power consumption of inference routines.
        Model optimization techniques for embedded deployment INT8 quantization, structured pruning, and knowledge distillation have seen significant advancements, particularly within the TensorFlow Lite Micro initiative and STMicroelectronics' STM32Cube. AI ecosystem or Edge Impulse. These tools enable the conversion of GPU-trained models into representations optimized for MCUs, with configurable accuracy, memory, and speed trade-offs. However, the validation of these approaches on real-world predictive maintenance scenarios involving noisy vibration signals, imbalanced classes, and temporal distribution shifts remains insufficiently documented.

    \subsection{Autoencoders and 1D CNNs for anomaly detection}
    
        Among the candidate model architectures for embedded anomaly detection, two families deserve special attention: autoencoders and 1D-CNNs. These two approaches offer complementary trade-offs that justify their combined use in a hybrid architecture.
        Unsupervised autoencoders learn a compressed representation (latent code) of the nominal signals from a healthy system. During the inference phase, reconstructing an input signal generates an error (MSE: Mean Squared Error) that serves as an anomaly signature: a healthy signal is reconstructed well (low MSE), while a degraded or faulty signal generates a high MSE exceeding an adaptive threshold. This approach has the advantage of not requiring labeled failure data, which is often scarce and costly to collect in industrial settings, and of detecting unexpected types of failures. The typical size of an autoencoder adapted to short-duration vibration signals (512 to 2048 samples) can be reduced to less than 50 KB in quantized parameters, making its deployment on STM32F4 technically feasible.
        1D CNNs, on the other hand, offer high-precision supervised classification capabilities, particularly well-suited for recognizing vibration patterns associated with identified failure modes (misalignment, imbalance, bearing defects, etc.). Their architecture convolutional layers, batch normalization, pooling, and dense layers can be scaled to fit within the memory constraints of industrial MCUs thanks to depthwise separable convolutions, which reduce the number of parameters by a factor of 8 to 10 without significant loss of accuracy. Recent studies \cite{LIS23, CHE23} have documented accuracies exceeding 95\% for multi-class detection of bearing faults using 1D-CNN architectures containing fewer than 100000 parameters.
        The hybrid autoencoder plus 1D-CNN architecture, in which the autoencoder serves as a sentinel (unsupervised anomaly detection) and the 1D-CNN serves as an expert (supervised classification of failure type), constitutes an original approach whose validation on embedded systems represents a significant scientific contribution. This escalation logic, from generic detection to precise classification, allows for the optimization of computational resources by activating the more computationally expensive model (1D-CNN) only when the autoencoder has identified abnormal behavior, thereby reducing the system's average computational load.

        \begin{table}[htbp]
        \centering
        \caption{Comparison of AI methods for embedded Digital Twins and MCU compatibility.}
        \label{table_4} 
        \resizebox{\textwidth}{!}{
            \begin{tabular}{l l l l l l l l}
                \toprule
                AI methods & Type & Flash(KB) & Inference latency(ms) & Precision & Labeled data & Novelty detection & MCU-compatible (STM32F4) \\ 
                \midrule
                Dense Autoencoder & unsupervised & 30--80 & 2--8 & anomaly detection & not required &  Excellent & ok \\
                CNN-1D compact & supervised & 5--150 & 5--20 & 88--97\% & required &  no & ok \\
                AE + CNN-1D hybrid & Hybrid & 80--200 & 7--28 & $>$85\% & partial &  excellent & ok \\
                standard LSTM & Sequence supervised & 200--800 & 20--100 & 85--95\% & required & no &  difficult \\
                Random Forest & supervised & 50--500 & 1--5 & 85--93\% & required & partial &  possible \\
                SVM (kernel RBF) & supervised & 100--400 & 5--50 & 82--95\% & required &  1-class detection &  difficult \\
                ResNet-1D & deep supervised & 500--2000 & 50--200 & 90--98\% & required &  no &  no \\
                Transformer-1D & supervised & 1000--5000 & 100--500 & 91--99\% & required &  no &  no \\
                Isolation forest & unsupervised & 100--300 & 3--15 & anomaly detection & not required &  excellent &  partial \\
                MobileNet-1D & supervised & 80--200 & 5--25 & 85--93\% & required &  no &  possible \\
                \bottomrule
            \end{tabular}
        }
        \end{table}

    \subsection{Estimation of Remaining Useful Life }

        Estimating the Remaining Useful Life (RUL) represents the most advanced stage of predictive maintenance, going beyond simple fault detection or classification to provide a time-based quantification of a component's progressive wear. This information can be directly applied to maintenance planning by replacing the conventional fixed-interval approach with condition-based maintenance and to the optimization of spare parts inventory.
        The correlation between the autoencoder's reconstruction error (MSE) and a component's degradation state has been documented in several studies \cite{ZHO23},\cite{CHE23},\cite{KOB24} which show that the MSE increases monotonically and in a relatively predictable manner during the progressive degradation of a bearing or gear subjected to a constant load. This property allows for the construction of an RUL estimator based on regression of the MSE trajectory, once a nominal degradation model has been characterized from historical data. LSTM (Long Short-Term Memory) models trained on MSE time series have demonstrated estimation errors of less than 10\% on several benchmark datasets (CMAPSS, PRONOSTIA/FEMTO).
        
        The work by Kobayashi and Alam \cite{KOB24} is particularly relevant in this context: it combines RUL estimation with explainable AI (XAI) techniques, enabling the model's predictions to be justified by identifying the frequency or time-domain features that contribute most to the estimated degradation. This aspect of explainability is crucial for the operational acceptance of automated predictive maintenance systems, in which operators and maintenance engineers must be able to verify and validate the system's recommendations before taking corrective action.
    
    
    \section{Technological challenges and open issues}

    \subsection{Interoperability and standardization}
    
        Interoperability between heterogeneous DTs is one of the most complex and least resolved challenges in the field. The proliferation of partially competing standards IEC's Asset Administration Shell (AAS), Microsoft Azure's DTDL (Digital Twins Definition Language), the OPC-UA Information Model, and Eclipse Ditto creates a fragmented landscape in which DTs from different vendors or designed at different times cannot easily exchange information. Klar et al. \cite{KLA24},\cite{AHS25} emphasize that the maturity of DT architectures is strongly correlated with their level of interoperability, which remains insufficient for most current industrial deployments.
        Standardization challenges intensify as one moves down to the embedded levels of the architecture. Industrial microcontrollers typically operate with real-time operating systems (RTOS): FreeRTOS, Zephyr, and ThreadX, which have proprietary software interfaces, complicating the integration of standardized communication libraries. The LwIP network stack, which serves as the reference implementation for TCP/IP on resource-constrained embedded systems, does not natively support certain security mechanisms required by industrial standards, necessitating additional abstraction layers that consume limited resources.
        The issue of communication security is another aspect of interoperability. Distributed DT architectures expose new attack surfaces, as documented by Azambuja et al. \cite{JAO24}, which include the manipulation of synchronization data between the physical and virtual models, the injection of false sensor data, and denial-of-service attacks on MQTT brokers. These vulnerabilities are particularly concerning in critical production environments, where a successful attack could result in physical damage to equipment or pose safety risks to operators.

    \subsection{Scaling and complexity management}
    
        Scaling up from a few demonstration nodes to hundreds or thousands of nodes in a real-world factory is a challenge that most academic research does not explicitly address. \cite{VRBA11} Architectures validated in the lab on 3 to 10 nodes often exhibit unexpected emergent behaviors when deployed at scale: MQTT broker saturation, increased decision propagation latency, conflicts between competing agents, and the accumulation of technical debt in aging DT models.
        MAS theory offers tools to address scalability, including hierarchical clustering mechanisms, authority delegation, and distributed consensus protocols. However, adapting these to the specific constraints of IIoT environments, limited bandwidth, variable latencies, unreliable communication links, and diverse hardware platforms remains an active area of research. Blockchain implementations and Byzantine Fault Tolerant (BFT) consensus protocols, although proposed in some works such as \cite{JON23}, incur computational overheads that are incompatible with current constrained embedded systems.
        Managing the life-cycle of DTs adds another layer of complexity. A DT is a living model that must be continuously updated to maintain its representational fidelity in the face of the gradual drift in the physical system's behavior (concept drift). In a distributed multi-agent architecture, this update must be orchestrated consistently across all levels of the hierarchy, without interrupting the monitoring service and without propagating modeling errors between levels.
    
    \subsection{Explainable Artificial Intelligence in the Digital Twins}
    
        The industrial adoption of deep learning-based predictive maintenance systems faces a fundamental methodological obstacle: the opacity of black-box models. Maintenance technicians and reliability engineers, trained to reason about comprehensible physical mechanisms, find it difficult to trust recommendations generated by neural networks whose decision-making process is imperceptible. This resistance to adoption is not irrational. It reflects a legitimate caution toward systems that can produce silent errors and whose validity limits are difficult to characterize.
        Explainable AI (XAI) techniques provide the methodological response to this challenge\cite{IQB26}. SHAP (SHapley Additive exPlanations), based on cooperative game theory, allows each input feature to be assigned a marginal contribution to the model's prediction. Grad-CAM (Gradient-weighted Class Activation Mapping), initially developed for images, has been adapted to one-dimensional temporal signals in the form of Grad-CAM 1D, enabling the identification of the time windows or frequency ranges that most influenced the CNN's decision. These two approaches are complementary: SHAP provides global explainability across all features, while Grad-CAM 1D offers fine-grained temporal localization of the discriminative information.
        Integrating these explainability mechanisms into digital twin architectures presents an additional challenge: SHAP calculations, which require repeated model evaluations with masked feature combinations, are computationally intensive and difficult to execute in real time on embedded systems. Fast approximations such as SHAP TreeExplainer for gradient boosting models or lightweight implementations of 1D Grad-CAM can be considered, but their explanatory accuracy and impact on operator confidence remain to be empirically validated. \cite{KOB24} is one of the few contributions that addresses this explainability-performance trade-off in the context of industrial DTs.

        \subsection{Security and safety: risk assessment}

        MAS- systems present a significantly broader attack surface compared to conventional SCADA or cloud-based monitoring systems, due to their distributed architecture, heterogeneous communication protocols, and autonomous decision-making capabilities \cite{AZA24},\cite{VOD24},\cite{RAV21}. The attack ecosystem can be organized into four categories: (i) attacks on data integrity-sensor spoofing and adversarial perturbations designed to mislead inference; (ii) attacks on availability-DDoS attacks targeting MQTT brokers or the communication channels of edge agents; (iii) model extraction attacks-reverse engineering of proprietary models through repeated inference requests; (iv) lateral movement-compromised edge agents used as pivot points for infiltrating the factory network.
        Saad et al. demonstrated that their microgrid MAS-DT could be deceived by sensor spoofing attacks that injected false frequency measurements, triggering erroneous load shedding decisions \cite{SAA20}. This class of attacks is particularly dangerous because it exploits the DTs trust in sensor data without requiring access to the AI model itself.  Contrary measures include anomaly detection (a meta-Sentinel monitoring the Sentinel's behavior), cryptographic attestation of sensors, and Byzantine-fault-tolerant MAS consensus protocols \cite{AZA24}.
    
        \subsection{Security and safety: embedded safety constraints}

        Implementing cryptographic security on resource-constrained microcontrollers imposes a significant computational overhead. Encrypting a 1 KB MQTT payload using AES-128 on the STM32F429I takes 0.8 ms when using the hardware AES accelerator, which is acceptable within the 50 ms real-time budget. However, the TLS 1.3 handshake for MQTT over TLS takes 180–400 ms due to certificate chain verification, which is prohibitive for real-time communication at 10 Hz update rates. Lightweight alternatives, including DTLS 1.3 (UDP-based TLS), pre-shared key (PSK) authentication, and MQTT over TLS with session resumption, reduce the handshake overhead to 15–40 ms, maintaining compliance with real-time constraints \cite{KHE22}.
        Physical security is an equally critical concern for agents deployed in the field: physical access to an embedded node enables firmware extraction, model theft, and configuration tampering. STM32 Read Protection (RDP) Level 2 provides irreversible flash memory protection but prevents debugging, a significant maintenance constraint. Hardware Security Modules (HSMs) such as the ATECC608B provide tamper-resistant key storage and attestation for microcontroller-based DT agents \cite{AZA24}.

        \subsection{Security and safety: confidence and repute}

        In an MAS-DT ecosystem where agents from different manufacturers or operators coexist (e.g., a DT installation incorporating machine twins from multiple suppliers), trust management becomes a critical coordination challenge. Hamzaoui and Julien introduced cyber-physical social DT networks in which agents maintain reputation scores for peer agents, downweighting the diagnostics of agents with histories of false alarms \cite{HAM22}. This approach draws inspiration from established MAS trust frameworks (FIRE, REGRET) but requires adaptation for resource-constrained platforms with limited memory for maintaining reputation histories.

    \section{Open research questions}
    
        The critical analysis conducted in the previous sections, combined with an examination of industrial needs and identified technological trajectories, allows us to formulate three open-ended research questions that serve as the guiding principles for a coherent research program on AI-driven DTs for industrial predictive maintenance.
    
    \subsection{Research question 1: embedding intelligence in constrained physical nodes}
    
        QR1: To what degree is it possible to design and deploy a hybrid AI agent architecture combining an autoencoder for anomaly detection and a 1D CNN for defect classification on an STM32F429I microcontroller, while simultaneously meeting the constraints of memory footprint ($< 350$ KB Flash), inference latency ($<$ 45 ms), and classification accuracy ($> 85$\%) characteristic of a viable industrial deployment?
        This question addresses a fundamental technological bottleneck: demonstrating that predictive intelligence can reside as close as possible to the monitored physical system, without relying on an external network infrastructure. Solving it requires original contributions in at least three areas: model architecture (design tailored to MCU constraints), embedded decision logic (sentinel-expert escalation mechanism), and validation on real industrial signals rather than standardized academic datasets.
    
        The success metrics for this research question are: (1) a multi-class classification accuracy of over 85\% on an industrial vibration dataset; (2) an inference latency measured on a real-world target of less than 45 ms; (3) a total footprint of the model and embedded logic of less than 350 KB of Flash memory; (4) a false alarm rate of less than 5\% over 30 days of continuous operation. These thresholds, established in accordance with the operational requirements of industrial maintenance, constitute concrete and measurable targets.
        
    \subsection{Research question 2: multi-node distributed coordination with low overhead}
    
        QR2: How can we design and validate an operational MAS running on STM32 nodes interconnected via MQTT/Nanopb, capable of maintaining effective coordination among agents including the detection of cascading failures and the rebalancing of monitoring responsibilities, while consuming less than 15\% of the available bandwidth on a typical industrial network (shared 100 Mbps Ethernet)?
        This research question addresses the architectural transition from a single embedded agent to an operational distributed MAS. It raises specific challenges in three areas. First, adapting the Contract Net Protocol to the real-time and memory constraints of microcontrollers, which requires simplifying the negotiation phases while preserving the desired coordination properties. Second, designing a compact message format (Nanopb vs. JSON) that minimizes bandwidth without compromising the semantic expressiveness necessary for coordination. Third, the definition and validation of MAS resilience metrics in the face of failures-agent node failure, partial loss of network connectivity, and gradual degradation of sensor performance.
        
        The experimental nature of this issue requires the construction of a multi-node test bed representative of real-world industrial topologies, with controlled fault injection to evaluate the MAS's behavior under degraded conditions. The expected experimental evaluation involves a systematic comparison between the distributed MAS   approach and a centralized reference approach, based on metrics such as latency, bandwidth, resilience, and detection accuracy.
    
    \subsection{Research question 3: hierarchical orchestration toward the Smart Factory}

        QR3: Can we define and validate a three-tier Hierarchical DT architecture component (STM32), machine (fog/Raspberry Pi), and plant (cloud) driven by a decentralized MAS, which simultaneously enables: (1) a RUL estimate with an error of less than 15\% over a 30-day prediction horizon; (2) hierarchical synchronization with an end-to-end latency of less than 500 ms; and (3) XAI explanations positively evaluated by industrial operators according to a standardized human evaluation protocol?
        This question represents the convergence of the two previous ones and addresses the highest ambition of the research program: the demonstration of a complete DT system, from the sensor to the factory dashboard, driven by multi-level distributed intelligence and capable of providing interpretable remaining useful life estimates. Resolving this requires original contributions on RUL estimation based on autoencoder MSE trajectories, on the MQTT-fog-cloud hierarchical synchronization architecture, and on the integration of SHAP or 1D Grad-CAM into an industrial supervision interface.
        Validating this question poses a particular methodological challenge: evaluating the quality of XAI explanations by human operators, which requires a controlled evaluation protocol (user studies, confidence questionnaires, A/B tests) whose rigorous design is in itself a methodological contribution. The relevance of the explanations for maintenance decision-making, reducing diagnostic time, improving the accuracy of the intervention, and reducing false positives constitutes the ultimate criterion for the system's acceptability.

    
    \section{Discussion}

    Industry 5.0 places human-machine collaboration, sustainability, and resilience at the center of industrial transformation, marking a break from the paradigm of maximum automation in Industry 4.0. Architectures driven by MAS must therefore balance sufficient decision-making autonomy to reduce the monitoring workload for operators with the transparency needed to enable human supervision.
    
    The comparative analysis conducted in this paper shows that the literature converges on a tripartite architecture: embedded agents at the component level, "fog" coordinators at the machine level, and "cloud" orchestrators at the factory level. We refer to this structure as the MAS-DT tripartite architecture. It has so far been documented only in fragments in the literature, but represents the logical culmination of ongoing developments.

    As part of this work, this hierarchical DTs architecture, driven by an MAS (MAS-DT), is currently being developed for predictive maintenance in Industry 5.0. 

            \begin{figure}[htpb]
                \centering
                \includegraphics[width=\textwidth]{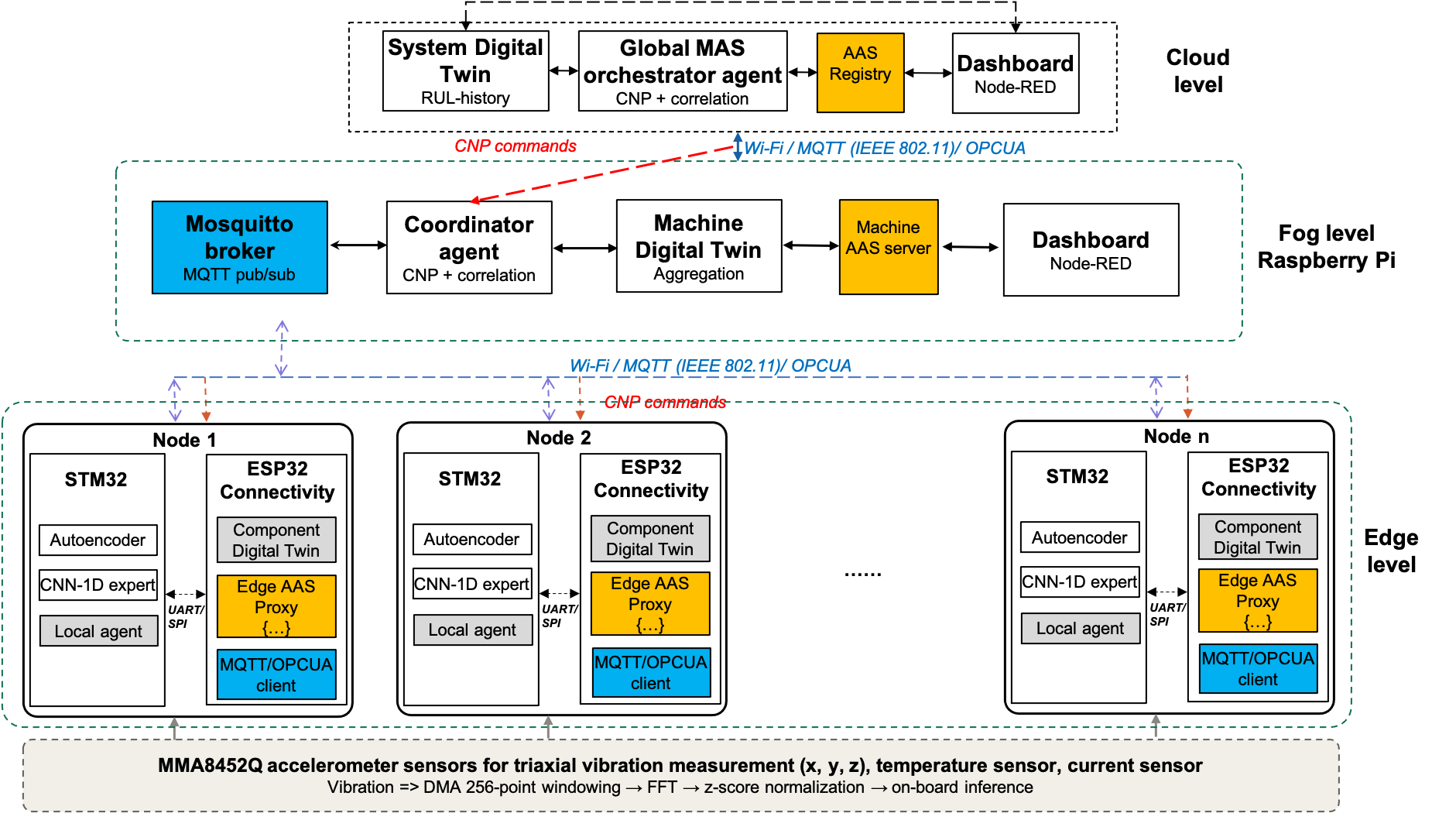}
                \caption{Our proposed MAS-DT tripartite architecture}\label{fig:tripatite_dt}
            \end{figure}
    
    Fig. \ref{fig:tripatite_dt} illustrates our proposed tripartite architecture. This architecture is based on a three-tier structure (Edge–Fog–Cloud), where each tier incorporates agents, DTs, and components of the AAS. AAS ensures interoperability between agents, DTs, and third-party systems (MES, ERP) through a standardized information model (IDTA). This is necessary for integrating the MAS-DT architecture into a heterogeneous industrial environment. At the Edge level, each component is equipped with an embedded node consisting of an STM32 and an ESP32: the STM32 handles data acquisition, the execution of embedded artificial intelligence models (autoencoder and 1D CNN), and local decision-making by an agent, while the ESP32 hosts a lightweight component DT, an AAS proxy, and communication services. At the Fog level, a coordinator agent aggregates information from the various component DTs to update a machine DT and expose the data via an AAS server. At the Cloud level, a MAS orchestrator manages interactions between multiple machines through a system DT and an AAS registry, providing a comprehensive view of the factory and supporting decision-making. Hardware and software development for this architecture, as well as its experimental validation on an industrial test bench, are currently in progress.

    This architecture contributes to the sustainability goals of Industry 5.0: by extending the service life of equipment, reducing excessive preventive maintenance, and optimizing the use of replacement parts, MAS-DT architectures reduce the material and energy footprint of production. Quantification of these benefits is still lacking in the literature and constitutes a complementary area of research.
    
    It also opens up a path toward the industrial metaverse \cite{REN24}, which integrates DTs into mixed-reality environments. Combining the decision-making capabilities of MAS systems with XAI interfaces in augmented reality could facilitate the operational acceptance of these systems, although current demonstrations remain at the prototype stage.
    A review of the articles highlights three systematic methodological weaknesses. 
    
    First, the reference data: 67\% of articles on fault diagnosis use the CWRU, which was recorded under controlled conditions with artificial defects and is not very representative of progressive degradation and composite failure modes encountered in the field. Validation using field data remains rare ($<15$\% of articles). 
    
    Second, the absence of ablation studies isolating the contribution of the MAS from that of the underlying AI model ($<30$\% of articles), which prevents determining whether the reported gains stem from multi-agent coordination or from the model’s architecture. 
    
    Third, the absence of energy analysis in the majority of articles (78\%), despite energy efficiency being cited as a motivation for edge deployment; in the absence of standardized benchmarks (µJ per inference), deployments are not comparable across platforms.

    \section{Conclusion}

        This systematic review examined the state of the art at the intersection of DTs and MAS, with a focus on predictive maintenance applications in resource-constrained environments. The analysis of over 547 high-impact scientific references enabled the establishment of a multidimensional taxonomy of existing architectures, the identification of persistent technological bottlenecks, and the formulation of three open research questions defining a coherent program of incremental scientific contributions.
        The main results of this analysis are as follows. First, MAS-driven architectures constitute a mature architectural response to the requirements of distribution, autonomy, and resilience in modern industrial maintenance systems, but their deployment on constrained microcontrollers remains insufficiently documented. Second, the autoencoder + 1D-CNN combination represents a promising hybrid AI architecture for on-board anomaly detection and fault classification, but its validation on industrial MCU targets and real-world signals remains a hurdle to overcome. Third, RUL estimation and explainable AI are key factors in the industrial acceptance of predictive maintenance systems, but their integration into a hierarchical DTs architecture has not yet been fully demonstrated in the literature.
        The research program, structured around the three research questions formulated AI deployment on MCUs, distributed MAS coordination, and hierarchical orchestration with RUL and XAI, defines a coherent path of scientific and technological innovation, in which each step contributes independently while fitting into a cumulative narrative toward the intelligent factory of Industry 5.0.

    \bibliographystyle{elsarticle-num} 
    \bibliography{cas-refs.bib}

\end{document}